\documentclass{article}
\usepackage{ijcai19}

\usepackage{times}
\usepackage{soul}
\usepackage{url}
\usepackage[utf8]{inputenc}
\usepackage[small]{caption}
\usepackage{graphicx}

\usepackage{float}

\usepackage{amsmath,amssymb} %
\usepackage{enumerate}
\usepackage{wrapfig}
\usepackage{multicol}  
\usepackage{multirow}
\usepackage{subfigure}
\usepackage{caption}
\usepackage{booktabs}
\usepackage{appendix}
\usepackage{pifont}
\usepackage[ruled]{algorithm2e}
\newcommand{\xmark}{\ding{55}}%

\title{Auto-MVCNN: Neural Architecture Search for Multi-view 3D Shape Recognition}
\author{
Zhaoqun Li$^1$
\and
Hongren Wang$^1$
\and
Jinxing Li$^1$
\affiliations
$^1$School of Science \& Engineering, The Chinese University of Hong Kong (Shenzhen)\\
}

\begin{document}

\maketitle
\begin{abstract}
    In 3D shape recognition, multi-view based methods leverage human's perspective to analyze 3D shapes and have achieved significant outcomes.
    Most existing research works in deep learning adopt handcrafted networks as backbones due to their high capacity of feature extraction, 
    and also benefit from ImageNet pretraining.
    However, whether these network architectures are suitable for 3D analysis or not remains unclear.
    In this paper, we propose a neural architecture search method named Auto-MVCNN 
    which is particularly designed for optimizing architecture in multi-view 3D shape recognition. 
    Auto-MVCNN extends gradient-based frameworks to process multi-view images, 
    by automatically searching the fusion cell to explore intrinsic correlation among view features. 
    Moreover, we develop an end-to-end scheme to enhance retrieval performance through the trade-off parameter search. 
    Extensive experimental results show that the searched architectures significantly outperform manually designed counterparts in various aspects, 
    and our method achieves state-of-the-art performance at the same time. 
\end{abstract}

%% ----------------------------
%% ----------------------------
\section{Introduction}
Along with the emergence of large 3D repositories~\cite{chang2015shapenet,wu20153d} and the development of Convolution Neural Network (CNN), 
deep learning based 3D shape recognition has attracted strong interest in research~\cite{su2015multi,xie2017deepshape,qi2017pointnet,su2018deeper,han2019seqviews}.
Among different kinds of research works, 
multi-view based methods have achieved the best performance so far, 
in which images are generally first rendered from a set of views and then passed into CNNs to obtain a shape descriptor.

Handcrafted networks are usually adopted as the backbone in current methods, 
where a variety of classic architectures (e.g. VGG~\cite{Simonyan15}, ResNet~\cite{He_2016_CVPR}) have been employed for feature extraction.
Over the past years, 
the majority of researches emphasized leveraging relationships among view images in single-view feature level~\cite{ma2018learning,He2019View} or multi-view feature level~\cite {feng2018gvcnn,han2019seqviews}, 
and therefore devoted efforts to designing sub-network on top of the backbone. 
Despite the remarkable progress achieved in previous studies, the effect of CNN extractors is not fully investigated, 
which restricts their performance to some extent.
Meanwhile, in order to avoid the excessive memory usage, 
a lot of research works~\cite{su2015multi,ma2018learning,He2019View} develop a multi-stage training scheme which only uses the backbone to extract view features, 
while the relation between the feature extraction and the feature fusion is neglected. 
These drawbacks may not only degrade the performance but also lead to being time-consuming and increasing computation cost.
As the neural network plays a crucial role in 3D shape recognition, 
it is desired to design an esfficient and powerful architecture that can process multi-view images with an end-to-end scheme.

In recent years, 
due to the effectiveness of Neural Architecture Search (NAS) compared with the human-designed structures~\cite{fang2019densely,guo2019meets},
its application field has also expanded on various benchmarks~\cite{zoph2018learning,real2019regularized,liu2019auto,Chen2019DetNAS}, 
before which NAS has been dedicated to image classification.
Besides, it is worth mentioning that the combination of multiple loss functions is essential in multi-task learning,
which has a large impact on neural architecture design. 
Unfortunately, many of AutoML methods using reinforcement learning and evolution algorithms have extreme
computational demands. And there is a relatively small amount of works that
study the balance of training as well as associated techniques in the searching
process.
Darts~\cite{liu2019darts} is a well-known gradient-based framework that largely reduces computation complexity. 
However, directly transferring Darts algorithm into 3D shape recognition is not an advisable choice. 
Firstly, Darts processes a single image instead of multi-view images, in which case the correlation across multiple views will be neglected. 
In addition, we aim to develop a unified model for both classification and retrieval tasks, 
which is different from Darts that focuses on single task optimization.

To automatically search a suitable neural network for 3D shape analysis, in this paper, 
we propose our Auto-MVCNN which is particularly adaptive for multi-view shape recognition.
Our network architecture contains three parts: 
a shared backbone for view feature extraction, 
a fusion module for multi-view feature fusion and a linear combination of loss functions for multi-task learning.
The pipeline of our method is shown in Fig. \ref{fig1}.
In the network, we propose a novel fusion cell which is specially designed for processing sequence view features.
By continuous relaxation of discrete operations, 
it inherits the efficiency and effectiveness of Darts and can be integrated into the existing framework seamlessly.
The equipped search space enables us to find appropriate fusion patterns that explore the correlation among views. 
For supervision signals, in addition to the shape classification loss on the top, 
we also add a view classification loss function for view feature enhancement and another retrieval loss function for the retrieval task.
The trade-off parameters that linearly combine these loss functions are searched by an end-to-end scheme.

To evaluate the performance of Auto-MVCNN, we carry out experiments on
two large-scale datasets and conduct the comparison from various aspects. 
Compared to the handcrafted network, with or without ImageNet pre-training, 
our searched networks show the superiority in regard to both performance and computation resources saving. 
Besides, more experiments are implemented to compare
our method with state-of-the-arts indicating the effectiveness of our proposed framework. 
Finally, we also analyze the impact of the number of initial channels and the stability of the
searching process.

To summarize, the contribution of our paper is four-fold:
\begin{enumerate}
  \item To our best knowledge, 
  this is the first work of neural architecture search in the field of multi-view 3D shape recognition that replaces the manually designed search with the automatic mechanism.
  \item We propose a novel fusion cell to process multi-view features that can be integrated into the existing framework seamlessly. 
  \item We develop a simple scheme that dynamically searches loss weights of multiple loss functions, 
  achieving appropriate training balance for multi-task learning.  
  \item Extensive experiments show that the searched CNNs achieve state-of-the-art performance, 
  using much fewer parameters than other baselines.
\end{enumerate}

%% ----------------------------
%% ----------------------------
\section{Related Work}
\subsection{Multi-view 3D Shape Recognition}
On the basis of different formats of the processed 3D data, 
methods in 3D shape recognition could be roughly divided into two categories:
model based methods~\cite{Osada:2002:SD,xie2017deepshape,li2019sgas} and view based methods~\cite{su2015multi,Wang2017DominantSC}. 
In this section, we mainly introduce multi-view based methods which leverage 2D views’ information to construct 3D descriptors.

MVCNN~\cite{su2015multi} is a typical framework in which the whole pipeline is divided into two parts.
The first part is the backbone for extracting view features and the other part is responsible for processing further shape features.
Between the two parts, the view features are aggregated into a single shape representation through the element-wise maximum operation.
In ~\cite{bai2016gift}, a postprocessing algorithm adopting the inverted file is proposed for fast retrieval.
Recently, leveraging correlation among views has become more and more popular in some research works.
GVCNN~\cite{feng2018gvcnn} introduces a multi-level descriptor by exploring the view-group-shape hierarchical correlation,
which largely improves the performance on 3D shape classification and retrieval.
~\cite{huang2018learning} develops a local 3D shape descriptor, which makes full use of relations over
points on the shape and can be directly utilized for a wide range of shape analysis tasks. 
Many research works~\cite{su2015multi,feng2018gvcnn,he2018triplet} show that metric learning is essential in 3D shape retrieval task.
\cite{li2019Rethinking} designs two loss functions to separately deal with these two distances.
The flexible combination property of the proposed loss functions provides effective tools to enhance retrieval performance.

% -----------------
\subsection{Neural Architecture Search}
The basic ideology of NAS is to find candidate network structures through a search strategy in a defined search space, based on the obtained feedback of the evaluation. 
The search space develops from the entire structure at the beginning to stacking cells~\cite{zoph2018learning}.
Cell-based search can greatly narrow the search space and improve the search efficiency, 
which has been applied in numerous subsequent works. 
However, search strategies based on reinforcement learning, 
such as the Q-learning algorithm in MetaQNN~\cite{BakerGNR17Desining},
require high computational complexity.
AmoebaNet~\cite{real2019regularized} develops evolutionary algorithms instead of reinforcement learning to optimize performance. 
Although it achieves better results, it still takes 450 GPUs and 7 days in a row to complete the experiment. 

NAS is gradually approaching the very obvious problem of solving heavy computation, 
which enables gradient-based methods and other efficient methods to emerge. 
ENAS~\cite{enas2018} employs weight sharing to accelerate validation, 
where the cell-based search mode greatly improves experimental results. 
Similar to ENAS, 
DARTS~\cite{liu2019darts} also searches subgraphs in designing cells and conducts weight sharing as well.
EfficientNet~\cite{Tan2019efficient} and MobileNetv3\cite{Howard_2019_ICCV} use the network search to obtain a fixed set of scaling factors to scale the width, depth, and resolution of the network respectively, 
achieving better efficiency and accuracy.  
To search for an appropriate loss function for face recognition,
AM-LFS~\cite{Li_2019_ICCV} employs the REINFORCE~\cite{Williams1992Simple} idea to automatically 
search for appropriate hyperparameters of the loss function, 
with great transferability at the same time.

%% ----------------------------
%% ----------------------------
\begin{figure*}[t]
  \centering
  \includegraphics[width=\textwidth]{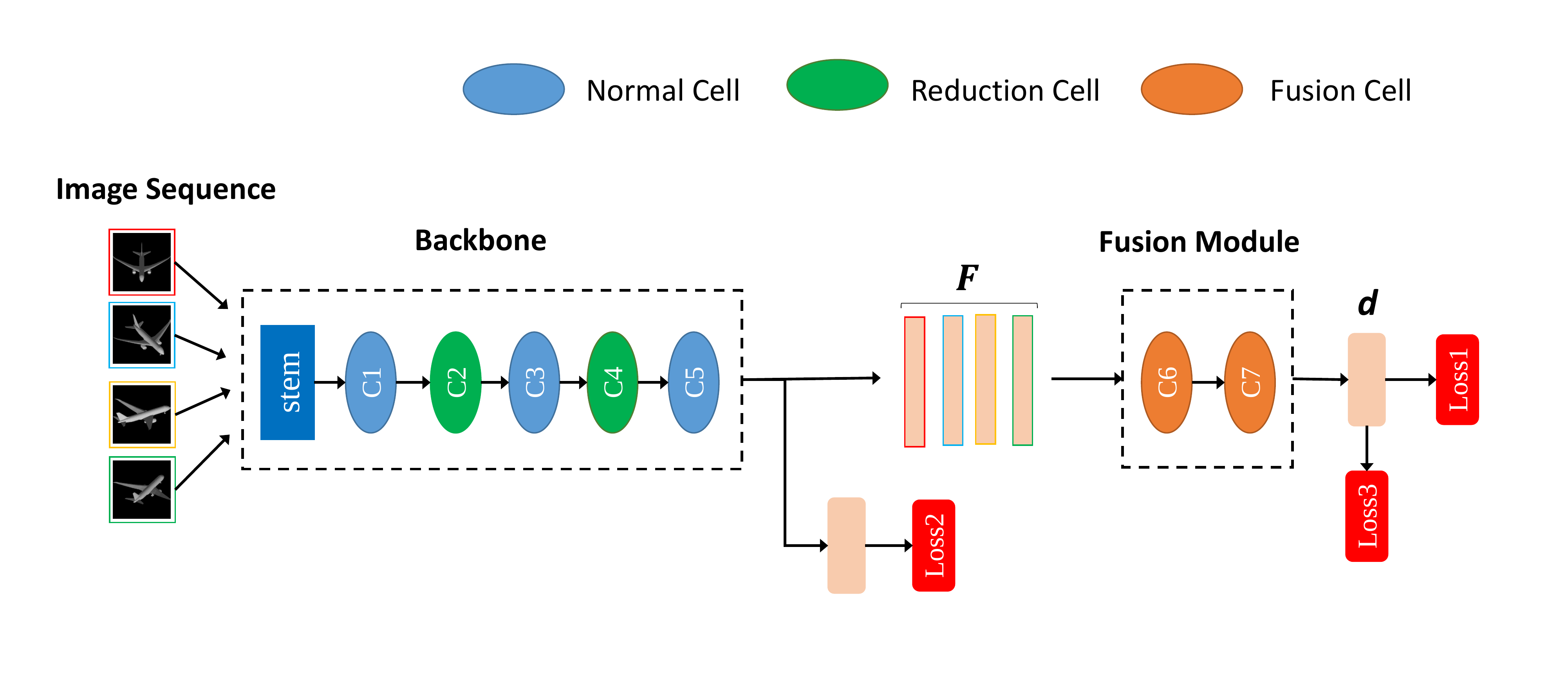}
  \caption{The illustration of Auto-MVCNN pipeline ($N_v = 4$). 
  The supernet is composed of backbone, fusion module and three loss functions.
  The stem in the backbone consists of several convolutional layers, each of which is a fixed structure.
  Three types of cells are learned in the searching process: normal, reduction and fusion.  
}
  \label{fig1} 
\end{figure*}

\section{Architecture Search}
The goal of Auto-MVCNN is to search for a neural architecture which is suitable for multi-view 3D shape feature representation.
In this section, we first describe the formation of our end-to-end pipeline.
Then we recall the definition of the neural cell and introduce our novel fusion cell for processing sequence features. 
Finally, we propose our loss combination scheme that automatically balances the training of classification and retrieval tasks.

% -----------------
\subsection{Auto-MVCNN Network} 
In our method, 
a view image sequence of length $N_v$ is input to a shared backbone to obtain view feature vectors 
$F = \left[ f_1, f_2,...,f_{N_v} \right] \in \mathbb{R}^{m \times N_v}$.
Then a shape descriptor $d \in \mathbb{R}^{n}$ is generated by fusing $F$.
The pipeline illustration is described in Fig. \ref{fig1}.

The whole neural network, which is called supernet, is formed of multiple cells and multi-task loss functions.
It is divided into two parts according to their functions: 
the backbone is used to extract view features and the fusion module is designed for view feature fusion. 
The backbone consists of a number of normal cells inserted with several reduction cells which are presented in a stacking manner.
Different from the backbone composition, the fusion module is generated by fusion cells. 
The details of these components will be described in Sec. \ref{sec_backbone} and Sec. \ref{sec_fusion}.

In our method, 
both classification and retrieval tasks are fulfilled with multiple supervision signals and 
we treat all the loss functions as important roles in neural network architecture.
Besides the classification loss $L_1$ on the top of the supernet, 
we add an auxiliary classification loss $L_2$ and an auxiliary retrieval loss $L_3$ to enhance view features and boost retrieval performance respectively. 
The formulation of the loss functions will be described in Sec. \ref{sec_loss}.

% -----------------
\subsection{Backbone Search} 
\label{sec_backbone}
In the searching stage, 
the cell is the basic searching component that contains the combination of all candidate operations.
Formally, a cell $C_{k}$ is defined as a directed acyclic graph (DAG) which contains an ordered sequence of 
7 nodes $x^{(1)}, x^{(2)}, \cdots,x^{(7)}$. 
Each node represents a latent tensor  (i.e. a feature map) and each edge $e_{i,j}$ consists of multiple parallel network layers.
Two input tensors $x^{(1)}$ and $x^{(2)}$ are the output of previous cells  $C_{k-1},  C_{k-2}$ 
and one output tensor $x^{(7)}$ is computed as $x^{(7)}=concat(x^{(3)}, x^{(4)},x^{(5)}, x^{(6)})$.
For the intermediate node, the computation can be formulated as
\begin{equation}
x^{(j)} = \sum_{i<j}\bar{o}^{(i,j)}(x^{(i)})
\end{equation}
where $\bar{o}^{(i,j)}$ is a continuous relaxation of the search space $\mathcal{O}$:
\begin{equation}
\label{equation_o}
\bar{o}^{(i,j)}(x)=\sum_{o \in \mathcal{O}} \frac{\exp(\alpha_{o}^{(i,j)})}{\sum_{o' \in \mathcal{O}}\exp(\alpha_{o'}^{(i,j)})} o(x)
\end{equation}
Here, $\alpha_{o}^{(i,j)}$ is the architecture parameter that represents the weight of operation $o$ in the edge $e_{i,j}$.
The continuous relaxation allows us to optimize $\alpha$ by gradient descent. 
$\mathcal{O}$ is the set of candidate operations and
we choose the same set as previous works~\cite{liu2019darts,zoph2018learning,Liu_2018_ECCV,liu2019auto} to keep consistence: 
$3\times3$ and $5\times5$ separable convolutions, $3\times3$ and $5\times5$ atrous convolution,
$3\times3$ average pooling,$3\times3$ max pooling, skip connection, and $\it{zero}$.

A cell that remains the same spatial resolution as the previous cell is the normal cell and that divides spatial dimension by 2 is the reduction cell.
In both training and inference stage, view features $F$ are extracted simultaneously from the backbone.

% -----------------
\subsection{Fusion Module Search}
\label{sec_fusion}

\begin{figure}
  \centering
  \includegraphics[width=0.9\linewidth]{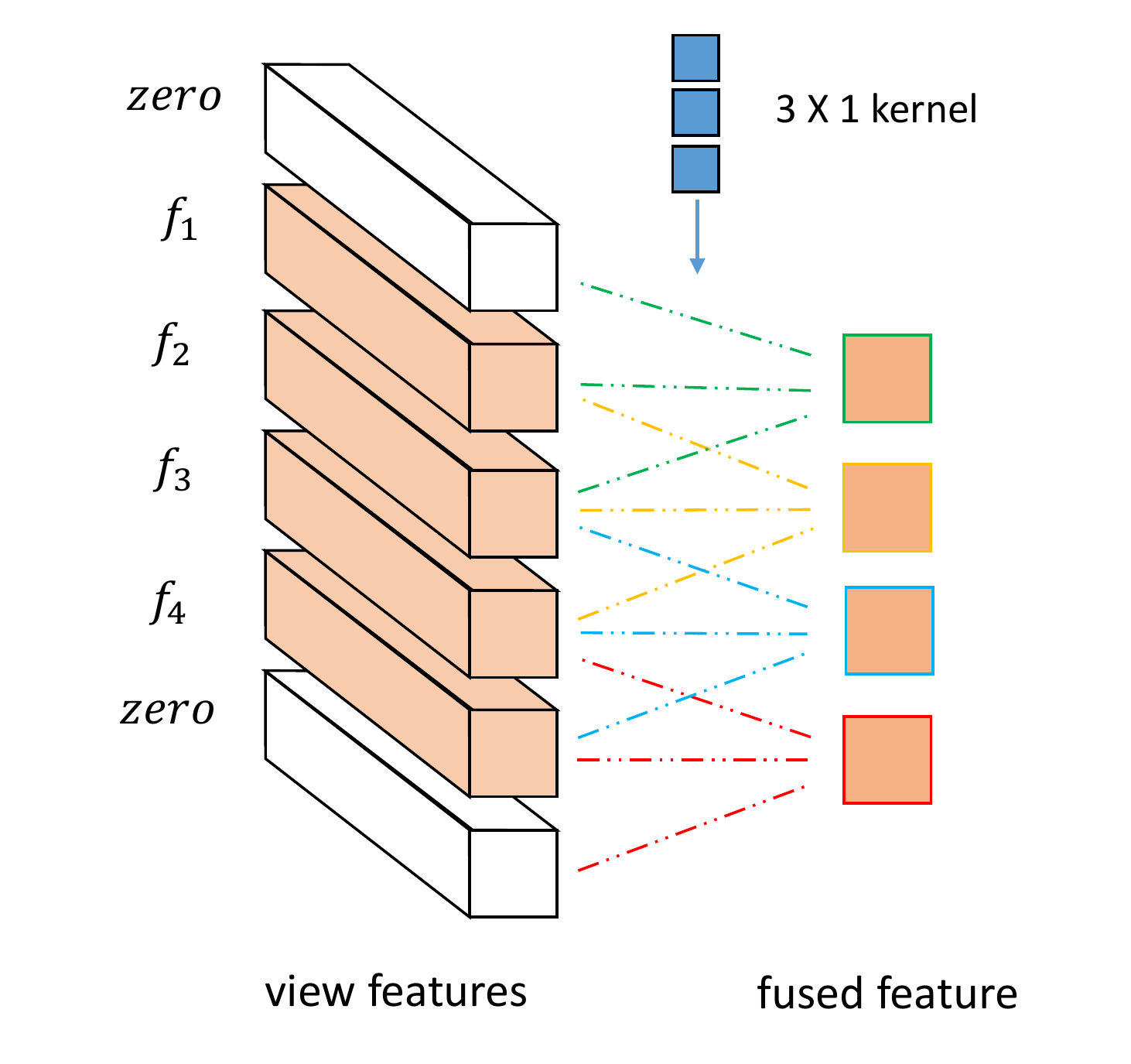}
  \caption{Fusion cell. We take $N_v=4$ and one $3 \times 1$ kernel for example. 
  Fused feature is obtained by iteratively aggregating 3 view features.}
  \label{fig2}
\end{figure}

The NAS framework~\cite{liu2019darts} focuses on image classification that can generate single view feature.
Though feasible, 
a simple combination (e.g. max-pooling, view-wise addition) of view-level features will lead to information loss that largely degrades the performance.  
By drawing on the experiences of previous works~\cite{feng2018gvcnn,He2019View}, 
we summarize that leveraging the spatial information and the feature correlation among views is essential for obtaining competitive performance.

Following the principle mentioned above, 
we design the fusion module in Auto-MVCNN to aggregate view features into a compact and discriminative shape feature.
This module consists of two sequential fusion cells that are developed to process sequence view features $F \in \mathbb{R}^{m \times N_v}$.
In order to integrate the fusion cell into the existing optimization framework,
the fusion cell has a search space $\mathcal{O}_f$ which is similar to $\mathcal{O}$ that applies the size adaption to the kernel size of all operations.
Concretely, 
we regard $F$ as a three dimension tensor of shape ${m \times N_v \times 1}$ with $m$ channels and spatial dimension $N_v \times 1$.
And the size adaption is to change the kernel size from $k \times k$ to $k \times 1$ for all operations in $\mathcal{O}$. 
In this way, we can adopt all the operations in $\mathcal{O}_f$ on $F$.
The size adoption of operations is illustrated in Fig. \ref{fig2}.
The fusion cell is then formed by linking these operations using Eq. \ref{equation_o}.
It is worth pointing that the size adaption is equivalent to padding zeros to $F$ such that its shape is ${m \times N_v \times k}$, 
while the operations remain the same in $\mathcal{O}$.

The compatibility of the fusion cell shows various excellent properties. 
For one operation $o \in \mathcal{O}_f$,  
$o$ conserves the spatial relationship of input tensors and thus can model the spatial information among views.
The correlation among view features can also be revealed by a variety of operations.
In addition, the fusion cell inherits the diversity of the combination among different layers, 
which enables the fusion module to search for novel fusion patterns. 

% -----------------
\subsection{Trade-off of Loss Functions}
\label{sec_loss}
Auto-MVCNN aims to develop a network for both shape classification task and shape retrieval task where training balance is extremely important.
There are totally three loss functions in the supernet. 
$L_1$ and $L_2$ are softmax loss located on the top and middle of Auto-MVCNN, 
in charge of the shape classification and the view feature enhancement respectively.
$L_3$ is the loss function proposed in~\cite{li2019Rethinking} which is used for enlarging the inter-class distance:
\begin{equation}
  L_3 = \sum_{i=1}^{M} \left[ \sum_{j=1,y_j \neq y_i}^{M}  \max (f_i \cdot f_j,0) \right]
  \label{equation_l3}
\end{equation}
where $M$ and $y$ are the batch size and the ground truth label respectively.

In general, classification places emphasis on the right label prediction while for retrieval the feature distribution is more important.
This phenomenon is also observed in~\cite{su2015multi,feng2018gvcnn} and they adopt offline metric learning algorithms to improve retrieval performance.
In this paper, by contrast, we tackle the issue by adding the loss function of metric learning $L_3$ with an end-to-end training scheme.
In practice, multi-task loss functions are linearly combined, $L = \sum_i \omega_iL_i$.
Since our target is searching for proper training balance, 
we normalize the loss weight using trade-off parameters $\lambda = (\lambda_1, \lambda_2, \lambda_3)$ and 
the total objective loss function is formulated as:
\begin{equation}
  L_{total} = \sum_{i=1}^{3} \left[ \frac{\exp(\lambda_i)}{\sum_{j=1}^{3}{\exp(\lambda_j)}} \cdot L_i \right]
  \label{equaion_l}
\end{equation}
An appropriate value distribution of $\lambda$ could enhance performance without impeding the classification task,
while a relative quick drop of loss $i$ means its gradient is large in the backpropagation which hampers the training of other tasks.
However, direct optimization via minimizing $L_{total}$ is infeasible.
If $L_i$ is the minimum value among the three losses, $\lambda_i$ will be straight to become large,
which is irrelevant to the balance of training.
To tackle this issue, in this paper, we develop a scheme that searches for the balance of training leveraging the performance on the validation set.
Specifically, let $L_i(t)$ denote the loss value in the $t$-$th$ iteration on the validation set,
and we define $r_i(t) = L_i(t-1)/L_i(t)$ as the training rate of $L_i$.
We propose a regularization term that directly involves $\lambda_i$ and the training rate: 
\begin{equation}
  L_{grad} = \sum_i^3\lambda_i(r_i - 1)
  \label{equation_lg}
\end{equation}
$L_{grad}$ penalizes $\lambda_i$ when its corresponding loss drops quickly and, in turn, 
it augments the weight of one task if its training is relatively slow.

%% ----------------------------
%% ----------------------------
\section{Training and Evaluation of Network}

\begin{algorithm}[t]
    \caption{The Auto-MVCNN search algorithm}
    \label{alg_1}
    \KwIn{architecture parameters $\left(\alpha, \lambda \right)$, network parameters $W$, $D_{train}$, $D_{val}$}
    \While{not converged}
    {
      Sample mini-batch from $D_{val}$, calculate $L_{total}(W, \alpha, \lambda)$ and $L_{grad}(W, \alpha, \lambda)$\;
      Update $\alpha$ by descending $\nabla_{\alpha} L_{total}(W, \alpha, \lambda)$\;
      Update $\lambda$ by descending $\nabla_{\lambda} L_{grad}(W, \alpha, \lambda)$\;
      Sample mini-batch from $D_{train}$, calculate $L_{total}(W, \alpha, \lambda)$\;
      Update $W$ by descending $\nabla_{W} L_{total}(W, \alpha, \lambda)$\;
    }
\end{algorithm}

\begin{figure*}
    \centering
    \includegraphics[width=0.9\textwidth]{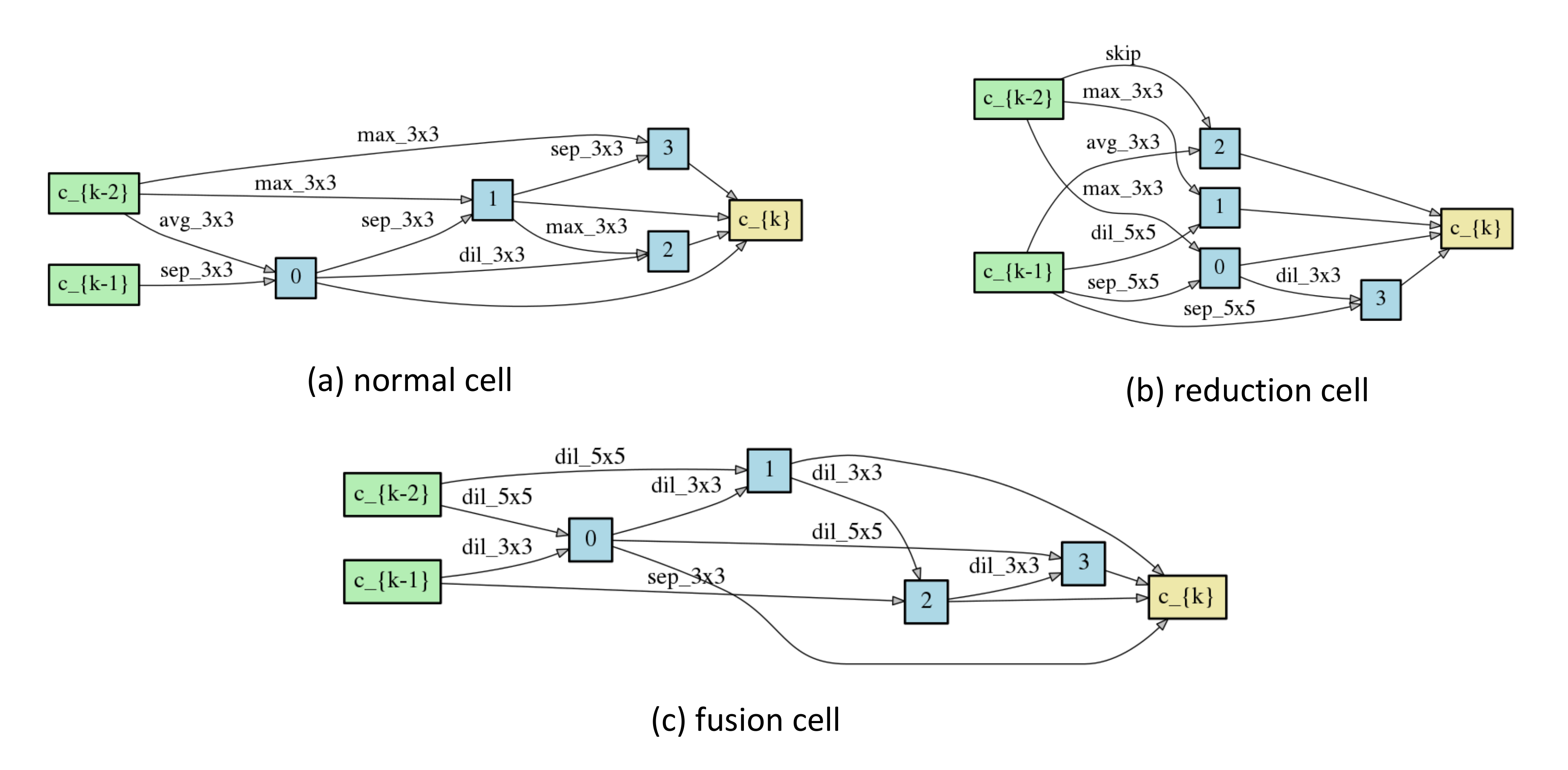}
    \caption{The learned cells on the ModelNet40. 
    The ``dil", ``sep", ``max", ``avg" in the figure represent depthwise-separable convolution, atrous convolution, 
    max-pooling and average-pooling respectively.}
    \label{fig3}
  \end{figure*}

\subsection{Optimization}
In our method, searching neural network is a bilevel optimization problem that has two different sets of parameters:
the network parameters $W$ and the architecture parameters $\left(\alpha, \lambda \right)$. 
We follow the first-order approximation proposed in Darts~\cite{liu2019darts} 
and split the training data manually into two disjoint sets $D_{train}$ and $D_{val}$.
The optimization of $W$ and $\left(\alpha, \lambda \right)$ is carried out in an alternating fashion on $D_{train}$ and $D_{val}$ until convergence, 
as shown in Alg. \ref{alg_1}.
The stability of search is clarified in Appendix.

% -----------------
\subsection{Evaluation}
\label{sec_eval}
After the search convergence, the final cell for evaluation is pruned by selecting non-zero layers in the connection.
The selection is achieved by retaining the top-k strongest operations from edge $i$ to edge $j$:
\begin{equation}
    \max_{k,o_k \neq zero} \alpha^{(i,j)}_{o_k}
\end{equation}
We use $k=2$ in our method. 
The extracted cells are then stacked to form the supernet and retrained for evaluation.
The final searched cells for evaluation are shown in Fig. \ref{fig3} and the final searched loss weights are $(0.216, 0.204, 0.580)$.
The search time is about 4 GPU hours.

For the sake of stability, the loss weights need to rescaled to ensure the loss weight of $L_1$ is 1, \textit{i.e.} $(1, 0.95, 2.7)$, in the retraining.
When we retrain the architecture, its initial number of channels changes to 24 and 36 (for matching the size of popular NAS architectures), 
generating our two representative models AM\_24 and AM\_36.

%% ----------------------------
%% ----------------------------
\section{Experimental Results}
\label{sec_exp}

\begin{table*}[t]
    \begin{center}
    \setlength{\tabcolsep}{1.2mm}{
    \scalebox{1}[1]{
    \begin{tabular}{lcccccccccc}
    \toprule[1pt]
    \multirow{2}{*}{Backbone} 	&& \multirow{2}{*}{Fusion layer} 	& \multirow{2}{*}{Params} 	& \multirow{2}{*}{MACs} & \multirow{2}{*}{Cputime}	 & \multicolumn{2}{c}{w/o pretrain}  & \multicolumn{2}{c}{w/ pretrain} 	 		\\
    \cmidrule(lr){7-8} \cmidrule(lr){9-10}       
    &&                                      & (M)		& (G)       & (ms)		    & Accuracy  & mAP & Accuracy  & mAP  \\		
    \toprule[1pt]
    VGG-M 		    && conv-5				  &	90.5	& 34.6	 	& 204           & 81.1\% 	& 66.2\%	& 90.4\% 	& 78.7\% 		\\	
    ResNet-18		  && resblock-5			&	11.2	& 21.9		& 111           & 84.5\%  	& 64.0\%	& 91.0\% 	& 81.7\%		\\
    ResNet-50		  && resblock-5			&	23.6	& 49.6		& 353           & 85.5\%  	& 63.4\%	& 91.3\% 	& 82.9\%		\\
    GoogleLeNet 	&& inception-5b		&	10.1	& 18.3		& 209           & 88.8\%  	& 75.0\%	& 91.9\% 	& 85.0\%		\\
    VGG-11		    && conv-6				  &	130.0	& 80.3	 	& 403           & 86.5\% 	& 80.2\% 	& 91.4\% 	& 86.1\%		\\
    VGG-11		    && conv-8				  &	130.0	& 90.5	 	& 460           & 86.9\% 	& 78.3\% 	& 91.3\% 	& 85.3\%		\\
    VGG-19 		    && conv-16				&	139.7	& 235.2	  & 970          & 82.4\%  	& 72.2\%	& 90.6\% 	& 86.6\%		\\
    \hline
    AM\_c24       && cell-5					&	2.1	  & 3.2		  & 200           & 90.5\% 	& 86.9\%	& 93.9\%  & 90.9\%		\\
    AM\_c36       && cell-5					&	4.7		& 6.9 	  & 289           & \textbf{91.0\%} & \textbf{87.6\%}	& \textbf{94.4\%}  & \textbf{91.0\%}\\
    \toprule[1pt]
    \end{tabular}
    }}
    \end{center}
    \caption{The performance comparison of different architectures on ModelNet40.}
    \label{tab_m40}
\end{table*}

% -----------------
\subsection{Dataset and Metrics}
To evaluate the performance of our method, 
we conduct experiments on the Princeton ModelNet dataset~\cite{wu20153d} and the ShapeNetCore55 dataset~\cite{savva2016shrec}. 

The ModelNet is a large-scale 3D shape dataset which contains 127,915 3D CAD models divided into 662 categories.
We apply the extracted subset ModelNet40, which includes 12,311 models cleaned manually from 40 categories, in our evaluation.
We follow the same training/testing split as described in~\cite{su2015multi}, 
by randomly selecting 100 unique models per category from the subset, 
where 80 models are used for training and the rest for testing.
The evaluation metrics adopted in this dataset include the (per-class) classification accuracy, 
the mean average precision (mAP) and the area under curve (AUC).
Their detailed definitions can be found in~\cite{wu20153d}. 

The ShapeNetCore55 dataset, 
introduced in the Shape Retrieval Contest (SHREC) 2016 competition track, 
contains 51,190 3D shapes from 55 common categories which is a subset of the full ShapeNet dataset with clean 3D models.
Each model in this dataset is attached with a label from the 55 categories plus a fine-grained subcategory deriving from 204 subcategories. 
The dataset is divided into two versions, 
named as “normal” version and “perturbed” version, 
where the 3D shapes in the former version are aligned but more challenging in the latter one with all shapes are rotated randomly. 
In terms of training and testing split method, 
70\% shapes in the dataset are provided for training and another 10\% shapes are for validation, 
with the remaining 20\% shapes forming the testing set.
Refer to~\cite{savva2016shrec} for the definition of metrics F-Measure (F-1) and NDCG used in this paper.

% -----------------
\subsection{Implementation Details}
\label{detail}
The experiments are carried out on a server with four Nvidia GTX2080Ti GPUs, Intel Xeon CPU E5-2678 v3 and 128G RAM. 
Before training and testing, each shape is rendered to generate 12 images with size 224$\times$224, 
following the same protocol as~\cite{su2018deeper}.

\noindent\textbf{Architecture search on ModelNet40.}
In the experimental settings, we employ 3 normal cells, 2 reduction cells and 2 fusion cells to build the architecture space.
The supernet contains a stem at the bottom with 7 cells stacked sequentially, Fig. \ref{fig1} displays the architecture. 
During the searching process, half of the ModelNet40 training data is set as the validation set $D_{val}$.
The batch size is 36 and the initial number of channels is 16.
Please refer to Appendix for other hyperparameter settings.

\noindent\textbf{Architecture evaluation on ModelNet40 and ShapeNetCore55.}
To evaluate the performance of searched architectures, 
we need to retrain the derived supernet on our target dataset. 
The retraining set the batch size to 36.
SGD with the initial learning rate 0.01 is adopted.
The shape descriptor $d$ is extracted for the retrieval task using cosine distance.
For a fair comparison with other methods, 
we also pretrain the supernet on ImageNet classification benchmark.
We propose two models, AM\_c24 and AM\_c36, with initial channels 24 and 36 respectively for evaluation. 
Please refer to Appendix for detailed hyperparameter settings.

% ----------------- 
\subsection{Main Results}

\begin{table}[h]
  \begin{center}
  \setlength{\tabcolsep}{1mm}{
  \scalebox{1}[1]{
  \begin{tabular}{lccccc}
  \toprule[1pt]
  Method          && Backbone			& Accuracy	& AUC		& mAP \\
  \toprule[1pt]
  3DShapeNet 			&& -			    	& 77.3\% 		& 49.9\%	& 49.2\%	\\	
  DeepPano 	      && -			    	& 82.5\% 		& 77.6\%	& 76.8\%	\\
  PointNet 	      && -			    	& 86.2\% 		& -			  & -	\\
  Octree		      && -			    	& 90.6\% 		& -			  & -	\\
  \hline
  MVCNN-su 		    && VGG-M		   		& 90.1\% 		& -			  & 80.2\%	\\
  ATCL 			      && VGG-M		   		& - 			  & 87.2\%  & 86.1\%	\\
  GVCNN 		      && GoogLeNet			& 93.1\% 		& -			  & 85.7\%	\\
  NCENet  	      && GoogLeNet			& -				  & 88.0\% 	& 87.1\%   \\
  MV-LSTM  	      && ResNet18				& 91.1\% 		& 85.7\%	& 84.3\%	\\
  RED 			      && ResNet50				& - 			  & 87.0\%	& 86.3\%	\\
  MVCNN-new   	  && VGG-11		  		& 92.4\% 		& -			  & -	\\
  TCL  			      && VGG-11		  	  & -  			  & 89.0\%	& 88.0\%	\\
  SeqViews        && VGG-19		  		& 91.4\%  	    & -			  & 89.1\%	\\
  VNN 			      && VGG-19		  		& -  			  & 90.2		& 89.3\%	\\
  \hline
  Auto-MVCNN 		                  && AM\_c24	  		& 93.9\% 	& 91.5\%  & 90.9\%	\\
  Auto-MVCNN 		                  && AM\_c36	  		& \textbf{94.4\%}  	& \textbf{91.6\%} 	& \textbf{91.0\%}	\\
  \toprule[1pt]
  \end{tabular}
  }}
  \end{center}
  \caption{The performance comparison with state-of-the-art methods on ModelNet40.}
  \label{tab_stoa}
\end{table}

\noindent\textbf{Comparison with hand-crafted networks.}
As for comparative experiments, we train several popular hand-crafted networks in this domain using the same training protocol.
The comparative experiments involve several aspects, 
results of which are indicated in Tab. \ref{tab_m40}.
We employ the network parameters (Params) and the multiply–accumulate operation (MACs) to measure the network size and the computation cost.
Cputime is the network inference time averaged by 10 times running. 
Their values are obtained by inputting an image sequence $(12 \times 3 \times 224 \times 224)$ into the network.
For a comprehensive comparison, we take the following factors into consideration:  
(1) the position of the fusion layer, (2) with or without ImageNet pretraining.
We can see that the two models with different sizes in our Auto-MVCNN, 
AM\_c24 and AM\_c36, apply the least amount of parameters and take up the lowest computation cost, 
consequently saving a lot of memory. 
In addition,
Auto-MVCNN obtains the highest value of both accuracy and mAP, 
demonstrating the best performance with or without ImageNet pretraining. 

\begin{table}
    \centering
    \setlength{\tabcolsep}{1mm}{
    \scalebox{0.9}[0.9]{
    \begin{tabular}{lcccccccc}
    \toprule[1pt]
    \multirow{2}{*}{Methods} && \multicolumn{3}{c}{Micro} & \multicolumn{3}{c}{Macro} \\
    \cmidrule(lr){3-5}  \cmidrule(lr){6-8}  && F-1 & mAP & NDCG & F-1 & mAP & NDCG &  \\
    \toprule[1pt]
    Wang  			        && 24.6\%  & 60.0\%  & 77.6\%    & 16.3\%  & 47.8\%  & 69.5\%    \\
    Li  				        && 53.4\%  & 74.9\%  & 86.5\%    & 18.2\%  & 57.9\%  & 76.7\% 		\\
    Kd-network          && 45.1\%  & 61.7\%  & 81.4\%    & 24.1\%  & 48.4\%  & 72.6\%		\\
    MVCNN 	            && 61.2\%  & 73.4\%  & 84.3\%    & 41.6\%  & 66.2\%  & 79.3\%    \\
    GIFT                && 66.1\%  & 81.1\%  & 88.9\%    & 42.3\%  & 73.0\%  & 84.3\%	  \\
    TCL                 && 67.9\%  & 84.0\%  & 89.5\%    & 43.9\%  & 78.3\%  & 86.9\%    \\
    VNN                 && 71.3\%  & 84.3\%  & 89.7\%    & 50.1\%  & 78.0\%  & 86.8\%    \\
    NCENet              && \textbf{73.3\%}  & 89.6\%  & 92.1\%    & 51.3\%  & 85.6\%  & 90.5\%    \\
    \hline
    Our    && 68.1\% & \textbf{91.1\%} & \textbf{92.3\%} 	& \textbf{51.4\%} & \textbf{86.2\%} & \textbf{91.2\%} 	\\
    \toprule[1pt]
    \end{tabular}
    }}
    \caption{The performance comparison on SHREC16 perturbed dataset.}
    \label{tab_shrec}
\end{table}

\noindent\textbf{Comparison with state-of-the-art methods.}
We choose model-based methods including 3DShapeNet~\cite{wu20153d}, DeepPano~\cite{shi2015deeppano}, PointNet~\cite{qi2017pointnet}, Octree~\cite{wang2017cnn},
and view-based methods including MVCNN-su~\cite{su2015multi}, MVCNN-new~\cite{su2018deeper}, ATCL~\cite{li2019angular}, NCENet~\cite{Xu2019Enhancing}, 
TCL~\cite{he2018triplet}, RED~\cite{Song2017Ensemble}, SeqViews~\cite{han2019seqviews}, GVCNN~\cite{feng2018gvcnn}, 
VNN~\cite{He2019View} and MV-LSTM~\cite{ma2018learning}  methods for comparison. 
The comparison results are indicated in Tab. \ref{tab_stoa} \footnote{We report per-class accuracy.} .
With ImageNet pretraining,
Auto-MVCNN has the greatest accuracy, AUC and mAP, in which the values achieve $94.4\%$, $91.3\%$ and $91.0\%$ respectively. 
Particularly, AM\_c36 achieves better performance compared to AM\_c24.

Our proposed Auto-MVCNN is also evaluated on ShapeNetCore55 perturbed dataset. 
This perturbed dataset is more challenging as all shapes are rotated randomly. 
Note that the architecture is also searched on the ModelNet40.
We choose the participants of the competition~\cite{3dor.20161092,klokov2017escape} and other popular methods as comparison.
As is shown in Tab. \ref{tab_shrec}, 
our method (AM\_c36) outperforms others in both mAP and NDCG metrics. 
We attribute the little lower performance on F-Measure to the transfer of datasets. 
% -----------------
\subsection{Ablation Study}

\noindent\textbf{Effect of fusion module.}
To demonstrate the effectiveness of our fusion cells, 
in our searched network, 
we manually replace the fusion cells with normal cells and conduct experiments on ModelNet40. 
Note that we maintain the same number of layers and same supervision signals.
We also choose three popular cell-based NAS networks that have similar network size to ours as comparison (the supervision is single softmax loss). 
As these architectures are searched for single image classification, 
we adopt a view max-pooling operation to fuse the view features after the penultimate layer as~\cite{su2015multi}.
As we can see from Tab. \ref{tab_fusion}, 
owing to the ability of the fusion module that can explore the intrinsic correlation among view features,
our learned network outperforms other NAS architectures.
When compared with \cite{Howard_2019_ICCV} and \cite{Tan2019efficient}, the following two other factors also contribute to performance improvement:
(1) Our network is derived directly from the ModelNet40 dataset while others are searched on classification benchmarks;
(2) Multiple supervision signals and corresponding appropriate loss weights enhance the performance on both shape classification and shape retrieval.

\noindent\textbf{Effect of dynamic loss balance.}
To reveal the superiority of our loss weights balance method, 
we choose several commonly adopted loss combinations and compare their performances.
The results are shown in Tab. \ref{tab_loss}. 
Note that values in the loss combinations need to be rescaled to conduct retraining (see Sec. \ref{detail}).
We can see from the first three experiments that both $L_2$ and $L_3$ are essential for competitive retrieval performance and our result is better than others.
When the loss combination is close to ours, its corresponding performance is also similar to ours.

\begin{table} 
    \begin{center}
    \setlength{\tabcolsep}{1.0mm}{
    \scalebox{1}[1]{
    \begin{tabular}{lccccc}
    \toprule[1pt]
    Architecture 	        & Fusion cell  & Accuracy & mAP \\
    \toprule[1pt]
    Darts\_v2		  		    & \xmark	     & 92.9\% 	  & 83.0\%	\\
    MobileNetv3  		      & \xmark	     & 92.1\% 	  & 73.3\% 		\\	
    EfficientNet\_b0 		  & \xmark	     & 90.3\% 	  & 71.6\%		\\
    \hline
    AM\_c24            		& \xmark       & 92.9\%     & 87.0\%    \\
    AM\_c24            		& \checkmark   & 93.9\%     & 90.9\%		\\
    AM\_c36            		& \xmark       & 92.8\%     & 88.8\%	  \\
    AM\_c36            		& \checkmark   & 94.4\%     & 91.0\%		\\
    \toprule[1pt]
    \end{tabular}
    }}
    \end{center}
    \caption{The performance comparison of different NAS architectures with ImageNet pretraining.}
    \label{tab_fusion}
\end{table}

\begin{table}
    \begin{center}
    \setlength{\tabcolsep}{1.0mm}{
    \scalebox{0.95}[0.95]{
    \begin{tabular}{lccccccc}
    \toprule[1pt]
    \multicolumn{3}{c}{Loss combination}    & \multicolumn{2}{c}{w/o pretrain}  & \multicolumn{2}{c}{w/ pretrain} 	 		\\
    \cmidrule(lr){1-3}  \cmidrule(lr){4-5} \cmidrule(lr){6-7}       		  
    $L_1$& $L_2$ &$L_3$             & Accuracy  	& mAP  	    & Accuracy  & mAP  \\		
    \toprule[1pt]
    1     & 0     & 0               & 88.8\% 		  & 77.4\%    & 90.5\% 		& 80.4\%	\\
    0.5   & 0.5   & 0 		          & 90.3\%      & 82.3\%    & 93.3\%    & 86.4\%  \\
    0.5   & 0     & 0.5		          & 88.6\%      & 82.2\%    & 89.9\%    & 83.8\% 	\\
    0.5   & 0.25  & 0.25 		        & 89.6\%      & 85.4\%    & 92.6\% 	  & 89.3\%  \\
    0.4   & 0.3   & 0.3	            & 87.4\%      & 82.8\%    & 93.5\%    & 89.9\%  \\
    0.2   & 0.2   & 0.6	 		        & 90.7\%      & 87.2\%    & 94.1\%    & 90.9\%	\\
    0.216 & 0.204 & 0.580		        & 91.0\%      & 87.6\%    & 94.4\%    & 91.0\%	\\
    \toprule[1pt]
    \end{tabular}
    }}
    \end{center}
    \caption{The results under different loss combination.}
    \label{tab_loss}
\end{table}

%% ----------------------------
%% ----------------------------
\section{Conclusions}
In this paper, aiming at the problem of multi-view 3D shape recognition,  
we propose a novel neural architecture search framework to optimize architectures, 
which is named as Auto-MVCNN. 
It abandons hand-crafted networks as the backbone for the first time, 
which greatly reduces the number of parameters and computational complexity.
The proposed fusion cell enables the whole network to explore the intrinsic connections of view features automatically, 
which fully utilizes the 3D information.
In addition, 
we apply a searching scheme for the training balance with an end-to-end fashion, improving both classification and retrieval performances. 
Extensive experiments exhibit our Auto-MVCNN achieves the best performance in various aspects,
and clarify its effectiveness at the same time. 

\bibliographystyle{named}
\bibliography{ijcai19}

\newpage

\appendix
\section{Training details}
\label{app_10}
\subsection{Architecture search on ModelNet40.}
\label{app_11}
We adopt the stochastic gradient descent (SGD) with the initial learning rate 0.01, 
the momentum 0.9 and the weight decay 3e-4 to optimize the network weights $W$.
The architecture parameter $\left(\alpha, \lambda \right)$ is initialized by a Gaussian distribution of the mean value 0 and the standard deviation 1e-3. 
$\lambda$ is optimized using the same optimizer of $W$ and the learning rate is 0.05.
$\alpha$ is optimized by Adam~\cite{kingma:adam} with the initial learning rate 3e-4, the momentum $(0.5, 0.999)$ and the weight decay 1e-3.
For the sake of stability, the gradient clip is adopted and a warmup scheme is conducted during the searching process.

\subsection{Pretrain on ImageNet}
\label{app_12}
In 3D shape recognition, most approaches take advantage of ImageNet pretrained network to boost their performances.
For a fair comparison, we also train the searched network on the ImageNet classification benchmark.
What should be noticed is that we train separate view images for the classification task, 
therefore only the parameters of the first 5 cells are updated.
The training process takes 800 as the batch size and total epochs are 120. 
SGD with the initial learning rate 0.1 and weight decay 3e-4 is used in the optimization.

\subsection{Architecture evaluation on ModelNet40 and ShapeNetCore55.}
\label{app_13}
To evaluate the performance of searched architectures, 
we need to retrain the derived supernet on our target dataset. 
The retraining set the batch size to 36.
SGD with the initial learning rate 0.01 is adopted.
The weight decay is 1e-3 without pretraining and 3e-4 with pretraining.
The shape descriptor $d$ with dimension $init\_channels \times 16$ is used to conduct retrieval.

\subsection{Stability of search.}
\label{app_22}
Since the searching process is initialization-sensitive, 
the final searched architectures are generally distinct from one another due todifferent random seeds. 
To investigate the performance stability of the searching process,
we conduct the search experiment 5 times with the same hyperparameters but different random seeds. 
The results of the experiments are shown in Tab. 6. 

\begin{table}
  \caption{The performance of searched architectures from different random seed on ModelNet40.}
  \begin{center}
  \setlength{\tabcolsep}{1.2mm}{
  \scalebox{1}[1]{
  \begin{tabular}{lccc}
  \toprule[1pt]
  Seed 			&& Accuracy		& mAP \\
  \toprule[1pt]
  1 		  		      && 90.3\% 		& 83.8\%	\\
  2			  	   	    && 91.0\%			& 87.6\%	\\
  3			  			    && 89.8\% 	  & 82.3\%	\\
  4	   	    		    && 90.5\%			& 84.2\%  \\
  5 	  				    && 90.5\% 		& 84.3\%	\\
  \toprule[1pt]
  \end{tabular}
  }}
  \end{center}
  \label{tab_seed}
\end{table}

\end{document}